\documentclass[11pt,a4paper]{article}
\usepackage[hyperref]{acl2021}
\usepackage{times}
\usepackage{latexsym}
\usepackage{amsmath}
\usepackage{array}

\newcolumntype{H}{>{\setbox0=\hbox\bgroup}c<{\egroup}@{}}

\usepackage[autostyle]{csquotes} 
\usepackage{todonotes}
\usepackage{multirow}

\usepackage{microtype}

\aclfinalcopy 
\def\aclpaperid{2585} 

\title{Weakly-Supervised Methods for Suicide Risk Assessment: \\Role of Related Domains}

\author{Chenghao Yang${^1}$,
  Yudong Zhang${^1}$, 
  Smaranda Muresan${^{1,2}}$ \\
  Department of Computer Science, Columbia University${^1}$ \\
  Data Science Institute, Columbia University${^2}$ \\
  \texttt{yangalan1996@gmail.com} \\
  \texttt{\{zhang.yudong, smara\}@columbia.edu}
  }

\date{}

\begin{document}
\maketitle
\begin{abstract}
 Social media has become a valuable resource for the study of suicidal ideation and the assessment of suicide risk. Among social media platforms, Reddit has emerged as the most promising one due to its anonymity and its focus on topic-based communities (subreddits) that can be indicative of someone's state of mind or interest regarding mental health disorders such as r/SuicideWatch, r/Anxiety, r/depression. A challenge for previous work on suicide risk assessment has been the small amount of labeled data. We propose an empirical investigation into several classes of weakly-supervised approaches, and show that using pseudo-labeling based on related issues around mental health (e.g., anxiety, depression) helps improve model performance for suicide risk assessment. 
  
\end{abstract}

\section{Introduction}
Suicide has been identified as one of the leading causes of deaths and approximately $1.5\%$ of people die by suicide every year \cite{worldsuicide,fazel2020suicide}. Despite years of clinical research on suicide, researoners have concluded that suicide cannot be  predicted using the standard clinical practice of asking patients about their suicidal thoughts \cite{mchugh2019association}. 
Recently, \citet{coppersmith2018natural} and \citet{nock2019advancing}  discuss the opportunities of using social media combined with natural language processing (NLP) techniques 
to complement traditional clinical records and help 
in suicide risk analysis and early suicide intervention.

To facilitate further research on automatic suicide risk assessment, \citet{zirikly2019clpsych} proposed a shared task, where they collected user data from r/SuicideWatch subreddit and annotated it with user-level suicide risk: no-risk, low-risk, medium-risk and high-risk. By comparing the results of the participating teams in this shared task, \citet{zirikly2019clpsych} conclude that one of the major challenges comes from the insufficient data for intermediate suicide risk levels (i.e., low risk and medium risk) rather than extreme risk levels (i.e., no risk and high risk). \citet{matero2019suicide} find that using a dual BERT-LSTM-Attention model to separately extract information from both SuicideWatch and Non-SuicideWatch posts together with feature engineering that includes emotion features, personality scores, user's anxiety and depression scores are important for model performance. 

In this paper, instead of feature engineering or complex model architectures, we explore whether weakly supervised methods and data augmentation techniques based on clinical psychology research can help improve model performance. 
We explore several weakly-supervised methods, and show that a simple approach based on insights from clinical psychology research \cite{o2014psychology} obtains the best performance. This model uses pseudo-labeling (PL) on data from the subreddits r/Anxiety and r/depression, which are considered important risk factors for suicide.  
We also present a potential application of our model for studying the suicide risk among people who use drugs, opening the door for using NLP methods to deepen our understanding between opioid use disorder (OUD) and mental health. 
 The code for this paper can be found at \url{https://github.com/yangalan123/WM-SRA}.
 
\section{Methods}
We focus on Task A from the CLPsych 2019 shared task ``Predicting the Degree of Suicide Risk in Reddit Posts"  \cite{zirikly2019clpsych}.
The goal of the task is to predict the user-level suicide risk category based on their posts in the r/SuicideWatch subreddit. Specifically, a user $u_{i}$ is associated with a collection of $n(i)$ posts  $C_i=\{x_{i, 1}, x_{i, 2}, \dots, x_{i, n(i)}\}$, where each post $x_{i, k} (1\leq k \leq n)$ has $m(i, k)$ sentences $x_{i, k} = [s_{ik, 1}, s_{ik, 2}, \dots, s_{ik, m(i, k)}]$. We need to predict $y_i \in \{a, b, c, d\}$ using $C_i$, where $a, b, c, d$ represent no-risk, low-risk, medium-risk and high-risk, respectively. 
In the original dataset, there are $496$ users in the training set and $125$ users in the test sets. We further split $100$ users from the training set to create the validation set. The sizes for the train/valid/test sets are 746, 173, and 186 respectively.   

\paragraph{Data Pre-processing}
Following the advice in \cite{zirikly2019clpsych}, we replace all human names and URLs in the Reddit posts with special tokens "\_PERSON\_" and "\_URL\_", respectively. We also remove punctuation and stop words besides lowercasing. Due to the limitation of GPU memory, we split those large posts to be passages with no more than $128$ words\footnote{The $128$ maximum passage length is tuned based on the validation set for both GPU memory and better computational efficiency for large posts. We do not observe a significant performance drop without a larger passage length.} and make sure that the split point is not in the middle of the sentence\footnote{We use a limited-size stack and greedily add each sentence into the stack. If adding a new sentence will make the sum of lengths of all sentences in the stack exceed $128$, we pop out all sentences, concatenate them to a new passage and then add this new sentence to the stack. For sentences having more than $128$ words, we treat them as individual posts.}. Such passages are treated as separate posts.

\paragraph{Model Architecture} Our architecture is a BERT \cite{devlin2019bert} model. We also experimented with other state-of-the-art pre-trained language models (PLMs), including RoBERTa~\cite{liu2019roberta} and XLNET~\cite{yang2019xlnet}, but found BERT to work the best and thus consider it as our baseline architecture (more details can be found in Appendix \ref{sec: comp_plms}). 
Each post $x_{i, k}$ 
is fed into BERT~\cite{devlin2019bert} and we get post embedding $\vec{e}_{i, k} = \text{BERT}(x_{i, k})$.
Then we do simple mean-pooling to obtain the user embedding $\vec{u}_{i} = \frac{\sum_{k=1}^{n(i)} \vec{e}_{i, k} }{n(i)}$.
Finally, we feed $\vec{u}_{i}$ to a fully-connected layer 
and use the Softmax layer to predict the risk level probability $\tilde{P}(y_i|C_i)$. The label with the largest probability is picked as the final prediction $\hat{y}_{i}$.
For training, the cross entropy loss $\mathcal{L}_{\text{clf}}$ is applied to optimize our model.

\subsection{Weakly-supervised Methods}
\paragraph{Task-Adaptive Pre-training}
%
Recent works \cite{lee2020biobert,gururangan2020don} point out that task-adaptive pre-training (TAP) can help pre-trained language models better adapt to the target domains and can bring improvement, especially in data-poor scenarios. Specifically, we continue pre-training (e.g., masked language modeling for BERT) on a task-relevant \textit{unlabeled corpus} and then do normal fine-tuning on the task. Our unlabeled corpus consists of all r/SuicideWatch posts (aggregated per user) from the training sets of all the tasks (A, B, C) in the shared task \cite{zirikly2019clpsych}. 
There are $621$ users and $138,057$ posts in this unlabeled corpus.
We do continued pre-training for $2$ to $3$ epochs and do early stopping.

\paragraph{Multi-view Learning}
Multi-view learning \cite{xu2013survey} (MVL) is one of the widely recognized semi-supervised methods. 
\citet{clark2018semi} provides a successful example of utilizing MVL in sequential labeling tasks. The idea is to create perturbations by masking words in certain positions and requiring the model to learn the similar distribution over the complete labeled examples and the corresponding masked examples besides normal classification training. However, since ours is a user-level classification task, we cannot directly borrow the same strategy from \cite{clark2018semi} as it mainly works on sequence labeling. We propose to create perturbations $\tilde{C}_i$ based on four strategies.\footnote{The masking proportions for \textbf{Word-mask} and \textbf{Sent-Mask} are tuned empirically on the validation set.}
First, for each sentence, we will randomly mask $10\%$ of tokens (\textbf{Word-Mask}). Second, considering that users may have posts of many words, we also propose a sentence-level masking strategy (\textbf{Sent-Mask}). For each post of a single user in the training set, we would randomly mask $10\%$ of tokens. Third, we only keep the beginning and ending sentences in each passage (\textbf{BegEd}). Usually these sentences convey the main purpose of the posts and should preserve  important semantics. Forth, we use Bert-extractive-summarizer 
\cite{miller2019leveraging} to extract the summary for each passage (\textbf{K-Sum}). It works mainly by first encoding each sentence $s_{ik, j}$ using a PLM to a continuous-valued representation $\vec{s}_{ik, j}$ and then training a K-means clustering over $\vec{s}_{ik, j}$. 
Finally it will pick $K$ sentences for each passage that are closest to the center. Empirically, we set $K=5$.
    

In training, we use KL-divergence to enforce the constraint that the predicted probability on perturbed examples $\tilde{P}(y_i| \tilde{C}_i)$ should be close to the one on complete examples (i.e., $\tilde{P}(y_i| {C}_i)$). The loss incurred by KL-divergence is simply added to the classification loss and these two losses are optimized together for each training instance.

\paragraph{Clinical Psychology Inspired Pseudo-labeling}
According to the analysis of the shared task report \cite{zirikly2019clpsych}, the main challenge for the 4-way classification comes from insufficient data for the intermediate classes (i.e., low-risk and medium-risk). A straightforward solution is to collect data for these two classes. Recent clinical psychological research \cite{o2014psychology} points out that mental health issues such as depression and anxiety can be important risk factors for suicide. Inspired by this study, we collect data from r/Anxiety 
and r/depression 
from Reddit. The time range of all collected data is from December 1, 2008 to September 30, 2020. 
We sample a small proportion of the collected data from both subreddits and after manual verification, we decided to assign \emph{low-risk} labels to all r/Anxiety users in the sample and \emph{medium-risk} labels to all r/depression users in the sample. Since we do not have experts to label these posts, adding too much pseudo-labeling data might introduce too much noise. 
Based on preliminary experiments on the \emph{validation set}, the number of added pseudo-labeling data is $8\%$ of the suicide risk assessment training data. The only difference between these experiments and the main experiments is that we only train the model for $10$ epochs rather than full $20$ epochs. Table \ref{tab:ratio_added_PL} show results for different sizes of added pseudo-labeled data from r/depression on the validation set. All pseudo-labeling data follows roughly the same pattern with the best proportion being $8\%$. 

\begin{table}[!h]
\centering
\small
\begin{tabular}{|c|c|}
\hline
$\frac{\#(\text{r/depression})}{\#(\text{Training})}$ & Macro-F1 on Validation set  \\ \hline
$2\%$ & $0.408$  \\ \hline
$8\%$ & $0.471$  \\ \hline
$16\%$ & $0.442$  \\ \hline
$32\%$ & $0.408$
\\ \hline
\end{tabular}
\caption{Results of different proportions of added pseudo-labeling data from r/depression.}
\label{tab:ratio_added_PL}
\end{table}

\begin{table}[t]
\centering
\small
\begin{tabular}{|c|c|c|HHHHc|}
\hline
No. & Approach & Setup & a & b & c & d & Macro (P/R/F1)\\ \hline
1 & Baseline & BERT & 0.742/0.719/0.730 & 0.077/0.077/0.077 & 0.400/0.286/0.333 & 0.525/0.615/0.566 & 0.436 / 0.424 / 0.427 \\ \hline
2 & TAP & BERT & 0.774/0.750/0.762 & 0.143/0.154/0.148 & 0.250/0.107/0.150 & 0.588/0.769/0.667 & 0.439 / 0.445 / 0.432 \\ \hline
3 & MVL & Word-Mask & 0.788/0.812/0.800 & 0.111/0.077/0.091 & 0.391/0.321/0.353 & 0.567/0.654/0.607 & 0.464 / 0.466 / 0.463 \\
4 & MVL & Sent-Mask & 0.551/0.844/0.667 & 0.091/0.077/0.083 & 0.294/0.179/0.222 & 0.583/0.538/0.560 & 0.380 / 0.409 / 0.383 \\ 
5 & MVL & BegEd  & 0.686/0.750/0.716 & 0/0/0 & 0.320/0.286/0.302 & 0.531/0.654/0.586 & 0.384 / 0.422 / 0.401\\
6 & MVL & K-Sum & 0.686/0.750/0.716 & 0/0/0 & 0.320/0.286/0.302 & 0.531/0.654/0.586 & 0.384 / 0.422 / 0.401\\
\hline
7 & PL & \begin{tabular}[c]{@{}c@{}}Depression
\\ 
(medium-risk)
\end{tabular}  
& 0.913/0.656/0.764 & 0.333/0.231/0.273 & 0.333/0.321/0.327 & 0.561/0.712/0.627 & \textbf{0.535} / \textbf{0.480} / \textbf{0.498}\\
8 & PL & \begin{tabular}[c]{@{}c@{}}Anxiety\\ 
(low-risk)
\end{tabular}  
& 0.808/0.656/0.724 & 0.167/0.154/0.160 & 0.440/0.393/0.415 & 0.565/0.673/0.614 & 0.495 / 0.469 / 0.478\\
9 & PL & \begin{tabular}[c]{@{}c@{}}Depression \\+ Anxiety
\end{tabular}  
& 0.821/0.719/0.767 & 0.133/0.154/0.143 & 0.385/0.357/0.370 & 0.554/0.596/0.574 & 0.473 / 0.456 / 0.463 \\

\hline
10 & PL & \begin{tabular}[c]{@{}c@{}}Task C\\ 
(low-risk)
\end{tabular} 
& 0.774/0.750/0.762 & 0.083/0.077/0.080 & 0.438/0.250/0.318 & 0.606/0.769/0.678 & 0.475 / 0.462 / 0.460 \\
11 & - & \begin{tabular}[c]{@{}c@{}}Task C\\ (crowd-\\labeled)\end{tabular} 
& 0.760/0.594/0.667 & 0/0/0 & 0.357/0.357/0.357 & 0.556/0.673/0.609 & 0.418 / 0.406 / 0.408 \\
 \hline

\end{tabular}
\caption{Results Task A test set. 
For each of tasks 7-11, the size of added data is 8\% of training data. Metrics are all reported on macro-average. 
}
\label{tab:exp-res}
\end{table}

\section{Experiments and Results}

We implement our BERT model based on huggingface Transformer \cite{wolf-etal-2020-transformers}. Due to the limitation of GPU memory, we only use the \textit{base} version.
We split $20\%$ of original training data to be the validation set and fix the split for all models. The model selection is made by early stopping and we train all models for $20$ epochs with the batch size $32$. For users with too many posts and words, we only sample $100$ passages for them. Table \ref{tab:exp-res} shows our results on Macro-F1.

\paragraph{Task-Adaptive Pre-training} After applying task-adaptive pre-training on BERT, we see small performance gains over BERT (i.e., from $0.427$ to $0.432$). That might be because even we use the whole corpus provided by the shared task, it is still not large enough. 

\paragraph{Multi-view Learning} 
 Word-Mask strategy improves over the BERT baseline. Compared with the adaptive pre-training results on BERT, which also do word-level masking but only trained on language modeling, we can see that MVL provides a more efficient way to utilize a small training corpus and bring $3.1\%$ gain on Macro-F1. However, all the other MVL approaches hurt the performance when compared to the BERT baseline.
This might be because the proposed sentence-level perturbation strategy can seriously break the semantics of each post and thus influence the overall performance, and random sampling over sentences hurts most. 

\begin{table}[t]
\centering
\small
\begin{tabular}{|HHc|c|c|c|cH|}
\hline
No. & Approach & Setup & a & b & c & d & Macro\\ \hline
1 & Baseline & Baseline & 0.730 & 0.077 & 0.333 & 0.566 & 0.427 \\ \hline
7 & PL & \begin{tabular}[c]{@{}c@{}}Depression
\\ 
(medium-risk)
\end{tabular}  
& 0.764 & 0.273 & 0.327 & 0.627 & 0.498\\

8 & PL & \begin{tabular}[c]{@{}c@{}}Anxiety\\ 
(low-risk)
\end{tabular}  
& 0.724 & 0.160 & 0.415 & 0.614 & 0.478\\

9 & PL & \begin{tabular}[c]{@{}c@{}}Depression \\+ Anxiety
\end{tabular}  
& 0.767 & 0.143 & 0.370 & 0.574 & 0.463 \\

\hline
10 & PL & \begin{tabular}[c]{@{}c@{}}Task C\\ 
(low-risk)
\end{tabular} 
& 0.762 & 0.080 & 0.318 & 0.678 & 0.460 \\
11 & - & \begin{tabular}[c]{@{}c@{}}Task C\\ (crowd-\\labeled)\end{tabular} 
& 0.667 & 0 & 0.357 & 0.609 & 0.408 \\
 \hline

\end{tabular}
\caption{Class-wise performance (F1) for PL-based methods (a=no-risk; b=low-risk; c=medium-risk; d=high-risk). 
}
\label{tab:PL-class-decomp-res}
\end{table}

\paragraph{Clinical Psychology Inspired Pseudo-labeling}
Exp 7, 8 and 9 in Table \ref{tab:exp-res} achieve the Top-3 Macro-F1 scores. This indicates that although our psychology-inspired pseudo-labeling technique is simpler than other weakly-supervised methods, adding meaningful pseudo-label data from relevant domains helps mitigate the problem of insufficient data in the intermediate classes (b and c).
 To verify this point, we show the class-wise classification results for PL-based models in Table \ref{tab:PL-class-decomp-res} where we can see improvements on b and c classes.  Due to space constraints, we present the class-wise performance for all models in Appendix \ref{sec:class-wise-decomp}. 

The investigation over the confusion matrix of the best model (shown in Section \ref{sec: human study})  further supports our hypothesis. 
However, when we try to combine different pseudo-labeling data together (see Exp 9, where we add users from r/depression and r/Anxiety following the proportion of $1:2$\footnote{See Supplemental material \ref{app: mixing_prop} for detailed experiments over different mixing proportions} 
and still keep the added user number the same), we observe a slight performance drop. The reason might be that users in these two PL datasets might be at the boundary of the low-risk and medium-risk and simply mixing them together will make the model confuse between  these two classes (see Supplemental material \ref{app:confusion_matrix} for all confusion matrices). 

\begin{figure*}[]
    \centering
    \begin{minipage}{0.48\textwidth}
        \includegraphics[width=\textwidth]{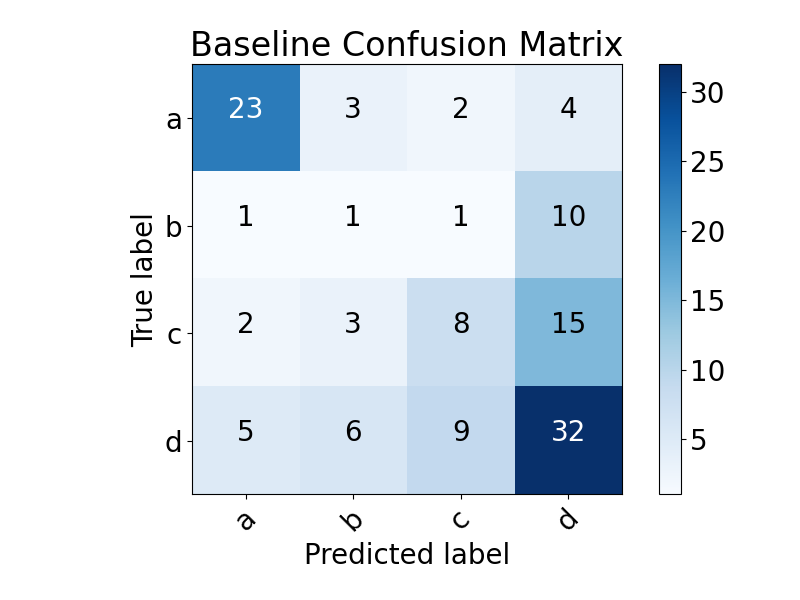}
    \end{minipage}
    \begin{minipage}{0.48\textwidth}
        \includegraphics[width=\textwidth]{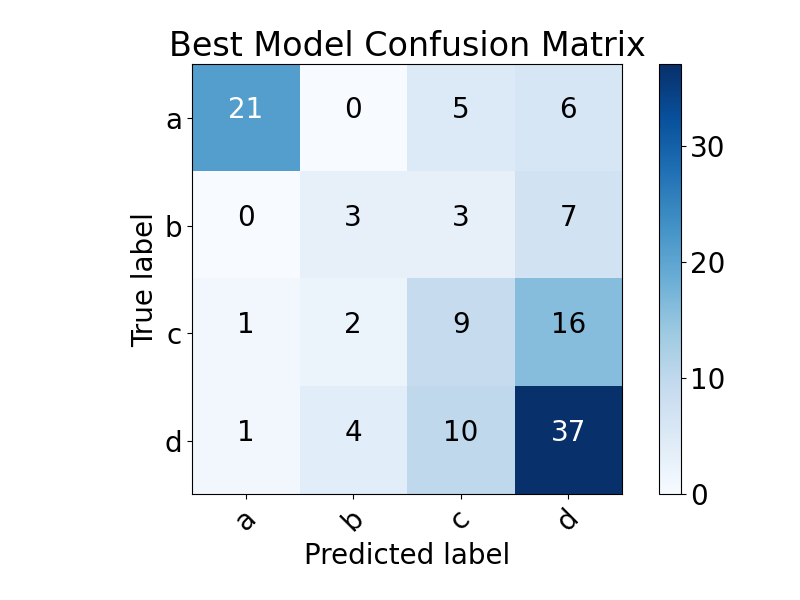}
    \end{minipage}
    \caption{Visualization of the confusion matrices for the baseline model (Exp 1)
    and the best model (Exp 7) 
    . 
    }
    \label{fig:Confusion_Matrix}
\end{figure*}

Furthermore, we wanted to test the role of the \emph{clinical psychology} aspect of our pseudo-labeling approach. Does the gain come from the meaningful domains (anxiety and depression) or just by adding additional data? 
To answer this, we use additional data provided by Task C of the shared task that contains posts from random subreddits (e.g., sports). We do two experiments: 1) assign low-risk to all such users and 2) assign the gold labels provided by the task via crowdsourcing. We add the same size as for the other pseudo-label experiment ($8\%$ of training data). 
The results (Exp 10 \& 11 in Table \ref{tab:exp-res}) show that the clinical psychology inspired PL outperforms these models by meaningfully addressing the intermediate classes insufficient data problem. 

\section{Error Analysis}
\label{sec: human study}
In this section, we take a closer look at the prediction results of our best model (clinical psychology inspired pseudo labeling using r/depression as medium risk) by looking at the confusion matrix and sampled error cases.
We plot the confusion matrices for the baseline model (Exp 1 in Table \ref{tab:exp-res}) and the best model (Exp 7 in Table \ref{tab:exp-res}) in Figure \ref{fig:Confusion_Matrix}. We can see that, the best model achieves the improvement mainly by fixing error cases wrongly predicted as no-risk (where the true labels are  ``b'', ``c'' and ``d'', with greater error reduction for "d") and low-risk (where the true labels are ``c'' and ``d''). As \citet{o2014psychology} point out, depression is a serious mental issue and has become  one of the most important risk factors of suicide. Adding posts from r/depression can help the model understand better what is ``medium-risk'' and ``high-risk'' and thus raise the alert for the signals of similar or related mental issues. 

We can also see that the main problem of our best model, is still the confusion between ``b'' (low-risk) and ``c'' (medium-risk). In addition, the problem of wrongly predicting the examples belonging to intermediate classes to high-risk ones still exists. By manual investigation, we find that both problems require expertise in mental health to make the subtle distinctions. 
For example, the following text comes from a low-risk example\footnote{Based on ethical consideration, we drop out many sensitive and private content of this example.} that is wrongly predicted as high-risk by our best model:
\begin{quote}
    ``
    \textit{
    \textcolor{red}{sadness has taken me}\dots
    \textcolor{red}{i am sad , lonely , and i have no interest in living anymore}\dots
    \textcolor{blue}{i didnt want to die}\dots
    \textcolor{red}{my mind is diseased , unable to take happiness}\dots 
    \textcolor{red}{i have no interest in forming any more.}
    }\dots
    \textcolor{blue}{i dont think ill do it}\dots
    ''
\end{quote}

It can be seen that there are many negative or even desperate expressions (marked as \textcolor{red}{red}) in this examples, mixed with some short signals (marked as \textcolor{blue}{blue}) possibly indicating a person 
considered at low-risk. 
The model can be fooled by the massive negative expressions and make the wrong predictions if the model is not aware of the true intent of the person.  Therefore, reliable intent identification that could consider user posts across time and other information would be a powerful tool to help the model prevent mistakes like this.

\section{Application: Predicting Suicide Risk of People Who Use Drugs}
In order to further verify the effectiveness of our model in real-world applications, we create a simulation scenario: we apply our best model (Exp 7) over the data that is collected for $612$ users who post on both r/opiates 
and r/SuicideWatch. r/opiates is a subreddit where people discuss topics around opioid usage (e.g., drug doses, withdrawal anguish, daily experiences, harm reduction).  
This community members could often be at a high suicide risk \cite{aladaug2018detecting, yao2020detection}. We apply our model over their  $1,176$ posts on r/SuicideWatch and find that our model predicts that $15.52\%$ of them are no-risk, while $84.48\%$ of them are of low-risk, medium-risk and high-risk. The results on sampled $2,863$ r/opiate posts are $30.56\%$ for no-risk and $69.44\%$ for at least some risk. The predicted outputs are highly aligned with reported results using crowdsourcing annotation of suicidal or not-suicidal by \citet{yao2020detection} and show the effectiveness of our model in this simulated scenario.\footnote{The original Mturk annotation dataset is not open-sourced and thus 
we can only do rough trend matching on our own collected data.} We hope this will open the door of using NLP methods to investigate the link between suicidal ideation and fatal overdoses among people who use drugs. 

\section{Conclusions}
We investigated a series of weakly-supervised methods and find that pseudo-labeling on data related to risk factors for suicide (depression, anxiety) can help improve model performance. This provides an alternative way to use theoretically-grounded models (e.g., compared to feature engineering). We also show a potential use case of this work for understanding suicidal ideation among users who use drugs (e.g., opiates).  

\section*{Ethical Considerations}
The dataset for suicide risk assessment was obtained from the organizers of the 2019 Clinical Psychology Shared Task on Suicide Risk Assessment, by filling in a participant application where we affirmed that we would follow the shared task’s rules. We have obtained IRB approval (exempt) from Columbia University  to  use the data as it consists of publicly available and anonymous posts extracted from Reddit.  For the application part, we also obtained Columbia IRB approval (exempt) for the data publicly available and anonymous data from r/opiates.  All data is kept secure and online userIDs are not associated to the posts. 

Our intention of developing and improving suicide risk assessment models is to help health professionals and/or social workers identify people that might be at risk of committing suicide. We emphasize our intention that suicide risk assessment models such as the ones developed here to be used responsibly, with a human in the loop --- for example a medical professional, a mental health specialist,  who can look at the predicted labels and offer explanations and decide whether or not they seem sensible.   We would urge any user of suicide risk assessment technology to carefully control who may use the system.  
Currently, the presented models may fail in two ways: they may either mislabel an at-risk user as no-risk (our current models are particularly designed to minimize this risk), or classify a no-risk user with some level of risk. Obviously, there is some potential harm to a person who is truly in need if a system based on this work fails to detect their suicidal ideation, and it is possible that a person who is not truly in need may be irritated or offended if someone reaches out to them because of a mistake. That is why, this system needs only to be used as additional help for health professionals. 

We note that because most of our data were collected from Reddit, a website with a known overall demographic skew (towards young, white, American men\footnote{\url{https://social.techjunkie.com/demographics-reddit}}), our conclusions about what expressions of different suicide risk levels look like and how to detect them cannot necessarily be applied to broader groups of people. This might be particularly acute for vulnerable populations such as people with opioid use disorder (OUD). We hope that this research stimulates more work by the research community to consider and model ways in which different groups express suicidal ideation.

\bibliographystyle{acl_natbib}
\bibliography{acl2021}

\newpage
\clearpage
\appendix

\section{Comparison of Different Pre-trained Language Models}
\label{sec: comp_plms}
Given that there has been significant progress on the architecture designs after BERT, we have experimented with different PLMs, such as RoBERTa \cite{liu2019roberta} and XLNet \cite{yang2019xlnet}. From Table \ref{tab:exp_plms}, we can see that on the Test set, the Macro-F1 scores for BERT and RoBERTa are almost the same and XLNet performs worse than BERT. 
Therefore, we hypothesis that the architecture of PLMs will not influence substantially the results on this task so we chose BERT model. 

\begin{table}[!htbp]
\centering
\small
\begin{tabular}{|Hc|c|c|c|Hc|}
\hline
No. & PLM & TAP? & PL? & MVL? & Valid & Macro-F1 \\ \hline
1 & BERT & No & No & No & 0.449/0.752/0.797 &
0.427
\\ \hline
2 & XLNET & No & No & No & 0.469/0.723/0.750 &
0.422
\\ \hline
3 & RoBERTa & No & No & No & 0.463/0.729/0.782 &
0.408
\\ \hline
\end{tabular}
\caption{Experiment results for different PLMs. Here we only show the macro-F1 for the baseline model built on different PLMs.}
\label{tab:exp_plms}
\end{table}

\section{Results for Different Mixing Proportions}
\label{app: mixing_prop}
Table \ref{tab:exp_mvl} shows the results for different mixing proportions of pseudo-labeling data from r/Anxiety and r/depression. Due to the limitation of space, in the main paper, we only show the results achieved by the best mixing proportions.
\begin{table}[h!]
\centering
\small
\begin{tabular}{|Hc|HHHHc|}
\hline
No. & Mixing Proportion & TAP? & PL? & MVL? & Valid & Macro-F1 \\ \hline
10 & 1: 5 & No & No & BegEd & 0.434/0.714/0.776 & 0.398 \\ \hline
4 & 1: 2 & No & No & K-Sum & 0.434/0.714/0.777 &  0.463 \\ \hline
3 & 1: 1 & No & No & Word-Mask & 0.486/0.746/0.797 & 0.434 \\ \hline
9 & 2: 1 & No & No & Sent-Mask & 0.430/0.733/0.777 & 0.441 \\ \hline
10 & 5: 1 & No & No & BegEd & 0.434/0.714/0.776 & 0.442
\\ \hline
\end{tabular}
\caption{Experiment results for different mixing proportions. Here the proportion represents the user ratio of $\#(\text{r/depression}) : \#(\text{r/Anxiety})$.}
\label{tab:exp_mvl}
\end{table}

\section{Class-wise Decomposition of Experimental Results}
\label{sec:class-wise-decomp}
Here we show the class-wise performance for all the models in Table \ref{tab:exp-class-decomp-res}.

\begin{table*}[]
\centering
\small
\begin{tabular}{|c|c|c|c|c|c|cH|}
\hline
No. & Approach & Setup & a & b & c & d & Macro\\ \hline
1 & Baseline & BERT & 0.742/0.719/0.730 & 0.077/0.077/0.077 & 0.400/0.286/0.333 & 0.525/0.615/0.566 & 0.436/0.424/0.427 \\ \hline
2 & TAP & BERT & 0.774/0.750/0.762 & 0.143/0.154/0.148 & 0.250/0.107/0.150 & 0.588/0.769/0.667 & 0.439/0.445/0.432 \\ \hline
3 & MVL & Word-Mask & 0.788/0.812/0.800 & 0.111/0.077/0.091 & 0.391/0.321/0.353 & 0.567/0.654/0.607 & 0.464/0.466/0.463 \\
4 & MVL & Sent-Mask & 0.551/0.844/0.667 & 0.091/0.077/0.083 & 0.294/0.179/0.222 & 0.583/0.538/0.560 & 0.380/0.409/0.383 \\ 
5 & MVL & BegEd  & 0.686/0.750/0.716 & 0/0/0 & 0.320/0.286/0.302 & 0.531/0.654/0.586 & 0.384/0.422/0.401\\
6 & MVL & K-Sum & 0.686/0.750/0.716 & 0/0/0 & 0.320/0.286/0.302 & 0.531/0.654/0.586 & 0.384/0.422/0.401\\
\hline
7 & PL & \begin{tabular}[c]{@{}c@{}}Depression
\\ (c) 
\end{tabular}  
& 0.913/0.656/0.764 & 0.333/0.231/0.273 & 0.333/0.321/0.327 & 0.561/0.712/0.627 & 0.535/0.480/0.498\\
8 & PL & \begin{tabular}[c]{@{}c@{}}Anxiety\\ 
(b)
\end{tabular}  
& 0.808/0.656/0.724 & 0.167/0.154/0.160 & 0.440/0.393/0.415 & 0.565/0.673/0.614 & 0.495/0.469/0.478\\
9 & PL & \begin{tabular}[c]{@{}c@{}}Depression \\+ Anxiety
\end{tabular}  
& 0.821/0.719/0.767 & 0.133/0.154/0.143 & 0.385/0.357/0.370 & 0.554/0.596/0.574 & 0.473/0.456/0.463 \\

\hline
10 & PL & \begin{tabular}[c]{@{}c@{}}Task C\\ 
(b)
\end{tabular} 
& 0.774/0.750/0.762 & 0.083/0.077/0.080 & 0.438/0.250/0.318 & 0.606/0.769/0.678 & 0.475/0.462/0.460 \\
11 & - & \begin{tabular}[c]{@{}c@{}}Task C\\ (crowd-\\labeled)\end{tabular} 
& 0.760/0.594/0.667 & 0/0/0 & 0.357/0.357/0.357 & 0.556/0.673/0.609 & 0.418/0.406/0.408 \\
 \hline

\end{tabular}

\caption{Class-wise decomposition results for models considered in this paper. The results under each class are presented following the  "Precision/Recall/F1" format.
}
\label{tab:exp-class-decomp-res}
\end{table*}

\section{Additional Error Analysis}
\label{app:confusion_matrix}
\begin{figure}[h!]
    \centering
    \begin{minipage}{0.48\textwidth}
        \includegraphics[width=\textwidth]{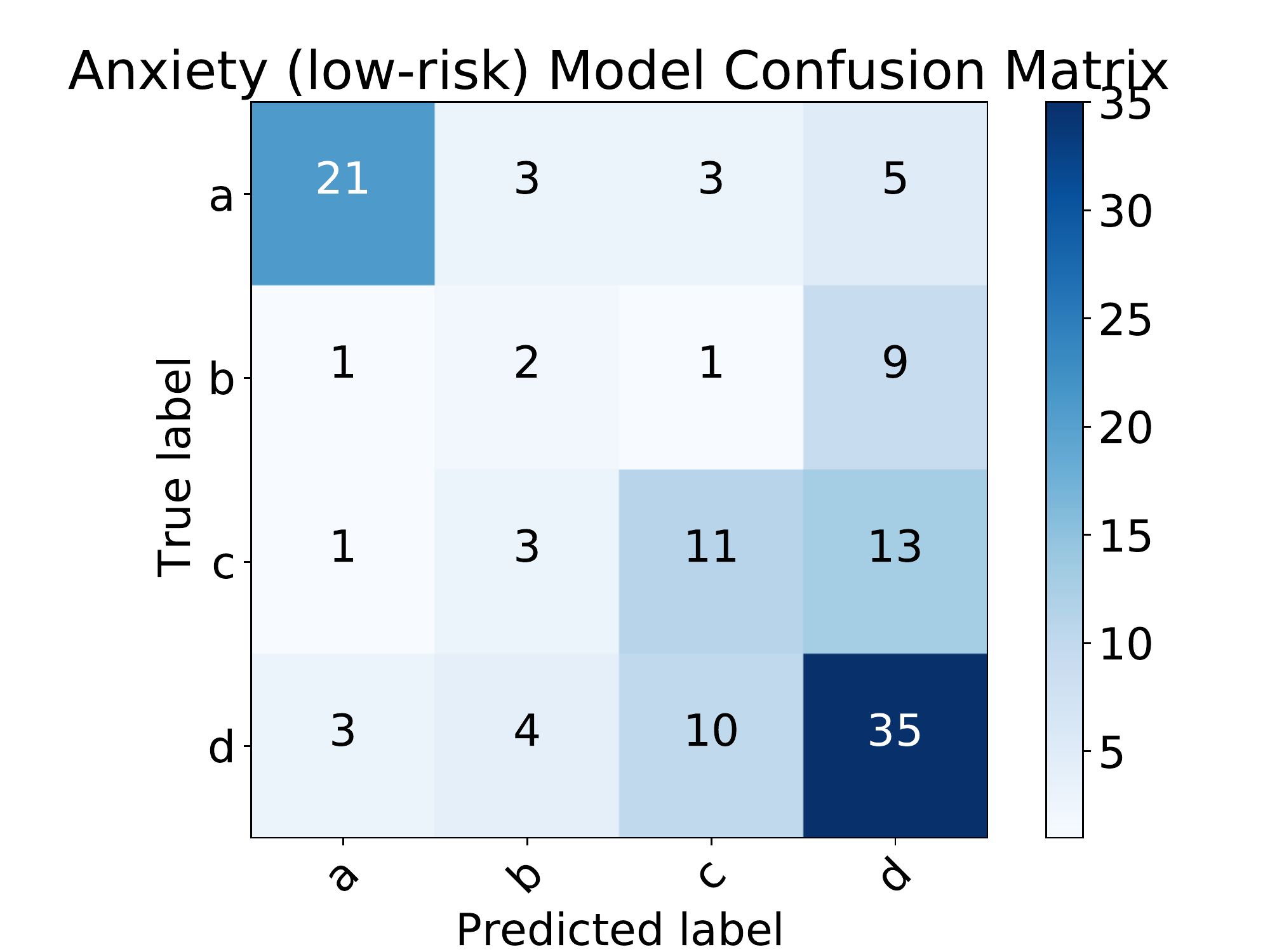}
    \end{minipage}
     \begin{minipage}{0.48\textwidth}
        \centering
        \includegraphics[width=\textwidth]{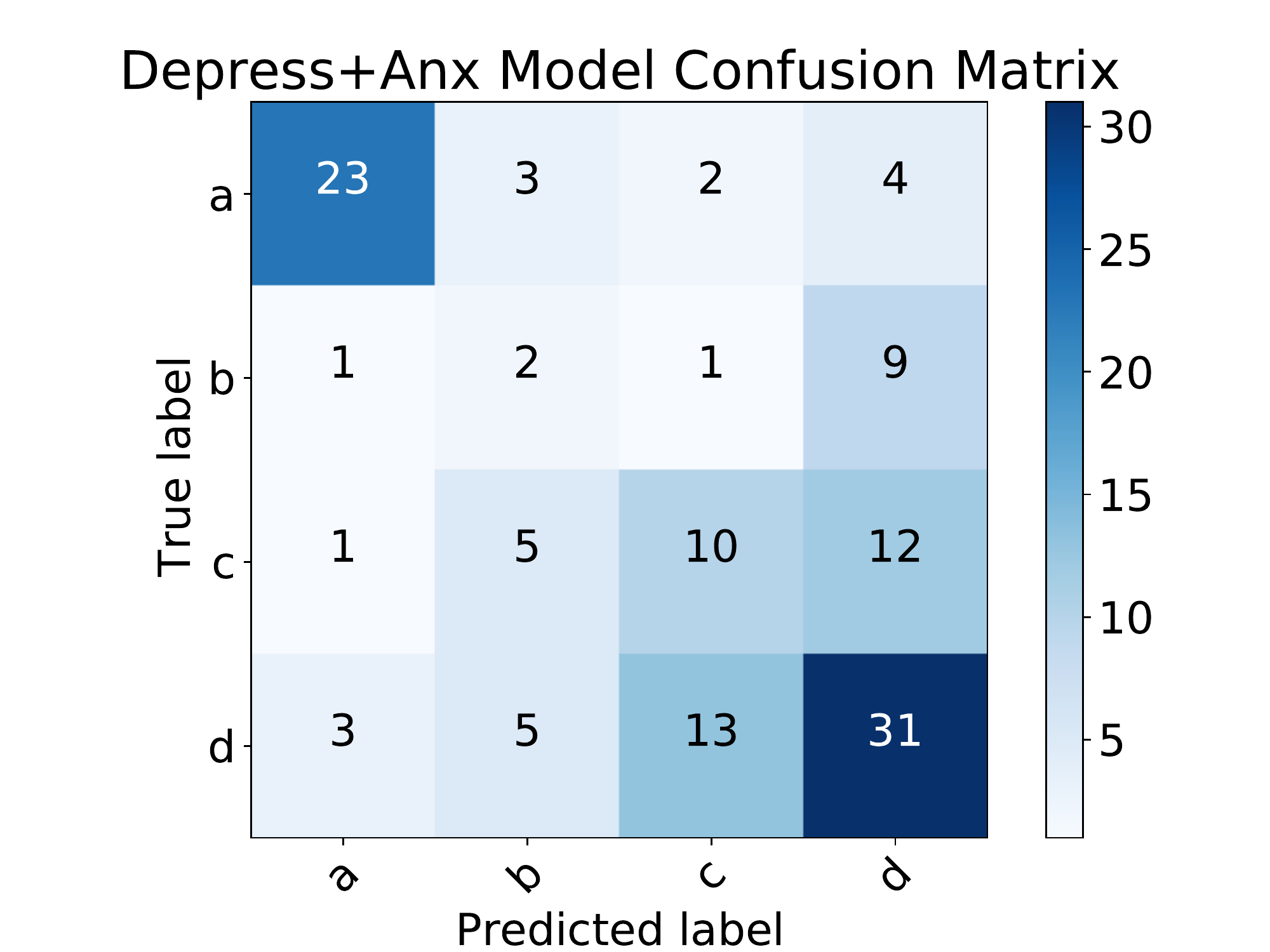}
    \end{minipage}
         \begin{minipage}{0.48\textwidth}
        \includegraphics[width=\textwidth]{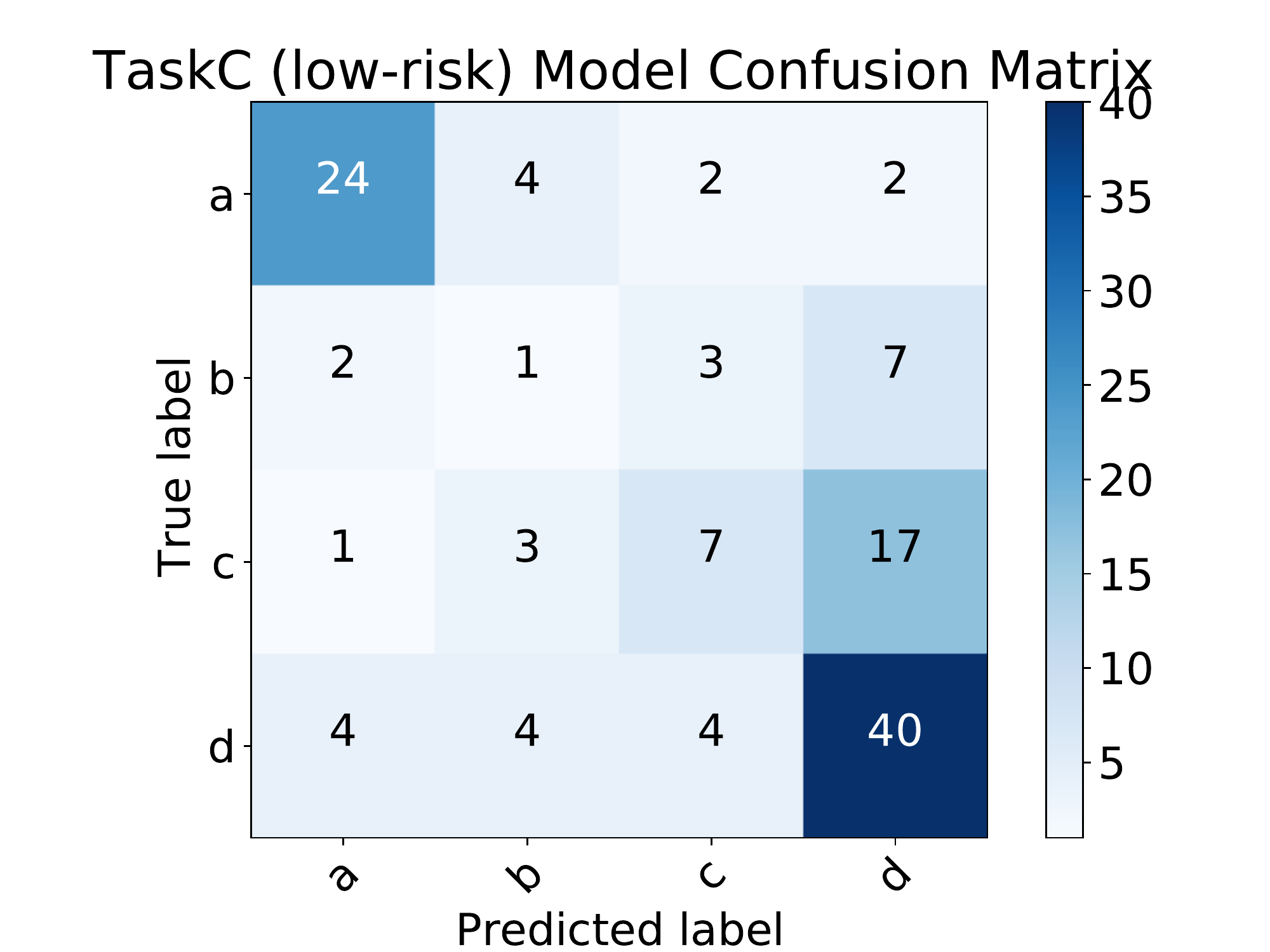}
    \end{minipage}
             \begin{minipage}{0.48\textwidth}
        \includegraphics[width=\textwidth]{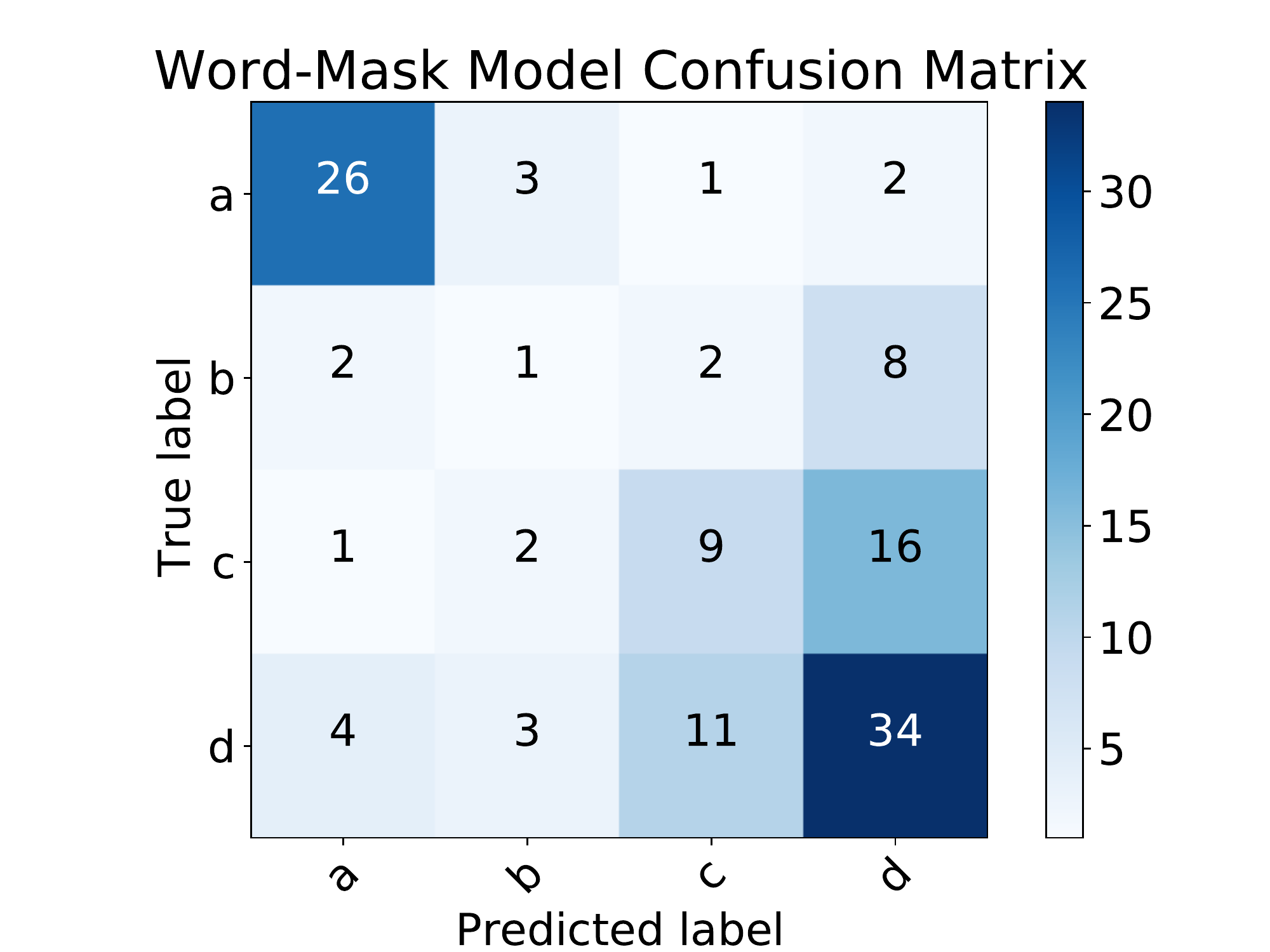}
        \caption{Word-Mask Confusion Matrix.}
    \end{minipage}
\caption{Additional Confusion Matrices for Task 8, 9, 10, 3 in Table \ref{tab:exp-res}}
\label{fig:addCM}
\end{figure}

Additional confusion matrices for high-performance models (8, 9, 10 in Table \ref{tab:exp-res}) are in Figure \ref{fig:addCM}.  
\end{document}